\documentclass[runningheads]{llncs}
\usepackage{graphicx}
\usepackage{amssymb}
\usepackage{amsmath}
\usepackage{siunitx}
\usepackage{subcaption}

\begin{document}
\title{Policy Entropy for Out-of-Distribution Classification}
%
%
\author{Andreas Sedlmeier \and Robert M\"uller \and Steffen Illium \and Claudia Linnhoff-Popien}
\authorrunning{Andreas Sedlmeier et al.}
%
\institute{LMU Munich, Munich, Germany\\
\email{andreas.sedlmeier@ifi.lmu.de}}
\maketitle              
\begin{abstract}
One critical prerequisite for the deployment of reinforcement learning systems in the real world is the ability to reliably detect situations on which the agent was not trained. Such situations could lead to potential safety risks when wrong predictions lead to the execution of harmful actions.
In this work, we propose PEOC, a new policy entropy based out-of-distribution classifier that reliably detects unencountered states in deep reinforcement learning.
It is based on using the entropy of an agent's policy as the classification score of a one-class classifier.
We evaluate our approach using a procedural environment generator.
Results show that PEOC is highly competitive against state-of-the-art one-class classification algorithms on the evaluated environments.
Furthermore, we present a structured process for benchmarking out-of-distribution classification in reinforcement learning.

\end{abstract}
\section{Introduction}

In the last years, impressive results were achieved using deep reinforcement learning techniques in areas as diverse as robotics or real time strategy games.
Despite these successes, systems built using these algorithms are still mostly deployed in controlled settings such as laboratory environments or video games.
Reliability of such learning systems when faced with changing observations in a real-world setting is still an open problem.
Being able to differentiate between states seen in training and non-encountered states can for example prevent silent and possibly critical failures of the learning system, caused by wrong predictions which lead to the execution of unfavorable actions.
We model the out-of-distribution (OOD) detection problem as a one-class classification problem, where only the in-distribution states are available at training time.
Having framed the problem this way, we propose PEOC, a new policy entropy based out-of-distribution classifier, which uses the policy entropy $H(\pi)$ as the classification score to detect OOD states.
PEOC classifiers can be constructed based on various RL algorithms from the policy-gradient and actor-critic classes.

\section{Preliminaries}

\subsection{Reinforcement Learning}

In the standard reinforcement learning (RL) formulation~\cite{sutton1998introduction}, an agent interacts with an environment defined as an MDP $\mathcal{M}$~\cite{puterman2014markov}
by executing a sequence of actions $a_t \in \mathcal{A}, t = 0, 1, ...$ over a possibly infinite number of discrete time steps $t$.
Each time step $t$, the agent is able to observe the state $s_t \in \mathcal{S}$ and to select an action $a_t \in \mathcal{A}$ according to it's policy $\pi$.
After executing the action, the next state $s_{t+1}$ is observed together with a scalar reward $r_t$.
The agents goal's is to find a policy $\pi : \mathcal{S} \rightarrow \mathcal{A}$, which maximizes the expectation of return $G_t$ at state $s_t$ over a potentially infinite horizon:
$R_{t} = \sum_{k=0}^{\infty} \gamma^{k} \cdot r_{t+k}$
where $\gamma \in [0,1]$ is the discount factor.

There are two fundamental approaches to reinforcement learning (RL), value based and policy based algorithms. In value based RL one seeks to find the optimal state-value or action-value function typically through an objective function based on the bellman equation. The optimal policy is then given by always selecting the action with the highest value in the current state. In contrast, policy-based methods directly search for the optimal policy $\pi^{*}(a|s)$ with parameters $\theta$. Modern deep RL approaches use neural networks to represent the policy or the value function and train by gradient descent on the parameters $\theta$.\\
Both approaches have their strengths and weaknesses. Value based methods can be trained off-policy where data from all previous runs can be reused for learning, leading to more sample efficiency. Policy based methods only use data from runs from the most recent version of the policy. However, they tend to be more stable than their value based counterparts.\\
Actor-critic algorithms are hybrid methods that combine both approaches \cite{mnih2016asynchronous,schulman2017proximal}.
The actor selects actions, while the critic is an estimate of the value function used to criticize the actor's actions.
In this work, we use proximal policy optimization (PPO)~\cite{schulman2017proximal}, an actor-critic algorithm that has
has proven successful in a wide variety of tasks ranging from robotics~\cite{andrychowicz2020learning} to real time strategy games~\cite{berner2019dota}.
Due to the non-stationarity of the training data (experience collected while acting according to the current policy), RL algorithms are often very unstable.
PPO aims to reduce this, by avoiding harsh changes in the policy's parameters. That is, the probability ratio 
\begin{equation}
    r_t(\theta) = \frac{\pi_{\theta}(a_t|s_t)}{\pi_{\theta_{old}}(a_t|s_t)}
\end{equation} 
between the old and the new policy, should not significantly differ from $r_t(\theta) = 1$.
In addition, an entropy bonus (based on the policy which can be interpreted as a distribution over actions) is often added to the loss function to refrain the agent from being overly confident and encouraging exploration. While PPO supports both discrete and continuous action spaces, the evaluations in this work focus on the discrete case. In this case, the policy's entropy with $n$ actions in some state $s_t$ can be readily computed by 
\begin{equation}
H(\pi(s_t)) = -\sum_{i=0}^{n} \pi(a_i|s_t) * \log \pi(a_i|s_t)
\end{equation} 

\subsection{Out-of-Distribution Detection}

Out-of-distribution (OOD) detection (also called novelty-, outlier- or anomaly-detection, depending on the specific setting and applied approach), is a thoroughly researched topic for low-dimensional settings.
The various approaches can be categorized as belonging to density-based, probabilistic, distance-based, reconstruction-based or information theoretic classes.
For an extensive survey on the topic of novelty detection with a focus on low-dimensional settings, see \cite{pimentel14}.
Further, it is important to differentiate between problems where samples from all classes are available at training time, versus problems where only samples of a single class are available.
The work at hand falls into the latter category (sometimes called one-class classification) as no OOD states (states that were not encountered during training) are available at training time.
While conventional out-of-distribution detection methods work reliably for low-dimensional settings, most break down with increasing dimensionality.
This is an effect of the curse of dimensionality, as in high dimensional input spaces, data becomes sparse and the real outliers become indistinguishable from noise being present in irrelevant dimensions.
For an in depth discussion of these effects and modern techniques, see e.g.~\cite{aggarwal2015outlier}.

With the rise of deep neural networks in the last years, new approaches were presented that try to tackle high-dimensional feature-spaces~\cite{hendrycks16,liang17}.
Still, most methods require access to training samples from all classes, and are not applicable to one-class classification problems.
One exception are deep autoencoder (AE) based approaches that try to learn useful representations of the training data in an unsupervised fashion.
AEs can be used to detect samples that were not part of the training data, by using the reconstruction error as the classification score~\cite{aggarwal2015outlier}.

\subsection{Evaluation of Binary Classifiers}\label{eval_binary_clfs}
As we model the OOD detection problem as a one-class classification problem, it is important to correctly evaluate the performance of different classifiers.
The basis of most approaches to evaluate the performance of a binary classifier is the classification score output by the classifier. 
Combined with a configurable threshold $t$, binary classification labels can be derived from these scores.
Consequently, the amount of true positives ($tp$) and false positives ($fp$) reported by a classifier, when applied to a dataset, depends on the chosen threshold.
A common choice to visualize this dependency is via a Receiver Operating Characteristic Curve (ROC).
It plots the true positive rate $tpr(t)$ on the y-axis against the false positive rate $fpr(t)$ on the x-axis (see Figure~\ref{fig:roc}).
When defining OOD samples as positives, the $tpr(t)$ (or recall) is defined as the percentage of ground-truth OOD samples correctly classified as OOD at threshold $t$.
The $fpr(t)$ then is the percentage of falsely reported positives out of the ground-truth negatives.
The ROC curve of a random classifier is the diagonal between $(0,0)$ and $(1,1)$,
while the curve of a classifier with better performance than random lies above this diagonal.
In addition to this visual evaluation, the area under the curve (ROC AUC) can be computed.
This is useful in order to compare the performance of different classifiers as well as repeated evaluation runs with different classifier configurations
(or in our case, different environment configurations and repeated RL training runs).

\section{Related Work}

\subsection{Out-of-distribution detection in deep RL}
Recently, an epistemic uncertainty based approach to detect out-of-distribution states was proposed in~\cite{ubood}.
The basic idea of the approach is that an agent’s epistemic uncertainty is reduced for in-distribution situations (states encountered during training), and thus lower than for unencountered (OOD) situations. The author's approach can be combined with different uncertainty estimation approaches like deep ensembles or Monte-Carlo dropout. The goal of this work is closely related to the work at hand, as it also tries to build a classifier to detect OOD states in deep reinforcement learning.
A limitation of the uncertainty based approach is that it is only applicable to value based reinforcement learning. Our proposed approach PEOC by difference is applicable to policy-gradient or actor-critic RL algorithms.

\subsection{Entropy regularization and maximum entropy RL}
The approach presented in this work differs in its goal from related reinforcement learning approaches dealing with policy entropy.
Most work considering policy entropy in RL is interested in using it during training, e.g. for exploration purposes during the learning phase.
As such, the probability distribution underlying the policy is used to introduce stochasticity in the action selection process.
One idea which can be categorized as entropy regularization initially proposed in~\cite{williams1991function} is to add the entropy of the policy $H(\pi)$ to the objective function in order to discourage premature convergence to local optima.
This idea was later successfully applied to various reinforcement learning algorithms~\cite{mnih2016asynchronous,schulman2017proximal}.
The extension of this idea, to not only find a policy with maximum entropy, but directly optimize the expectation of the entropy is called maximum entropy reinforcement learning~\cite{haarnoja2017reinforcement}.
It not only tries to optimize the policy entropy of visited states but also optimize the policy to seek out states that have high entropy.

Although we also focus on the policy entropy, the goal of our work is very different, as we are not trying to improve the learning performance of the RL algorithm.
The question considered by the work at hand is whether the policy entropy can be used to detect OOD states after the learning phase has completed.

\section{Policy Entropy for Out-of-Distribution Classification}

This section presents a new type of policy based out-of-distribution classifier that can be applied in deep reinforcement learning settings, we call PEOC (Policy Entropy Out-of-distribution Classifier).
We show how the policy entropy $H(\pi)$ of a RL agent can be used to detect OOD states.
PEOC classifiers can be constructed based on various RL algorithms from the policy-gradient and actor-critic classes.
These types of algorithms use a stochastic policy $\pi$ which is determined by a conditional probability distribution $p(a|s)$ defining the probability of taking action $a$ in state $s$ of the environment.
The policy entropy $H(\pi)$ then quantifies how random the actions being taken by an agent following the policy are.

The goal of RL is to maximize the expected future return, which is achieved by finding the optimal (state-dependent) action-sequences. 
Assuming that optimal behavior in most cases means acting non-randomly, the idea of PEOC then boils down to the hypothesis that the entropy of the action distribution has to decrease for states encountered during training in order to act optimally.
If this is the case, the policy entropy $H(\pi)$ can be used as the score of a binary classifier to detect OOD states.

Expressed more formally, a successful training process reduces $H(\pi(s_i))$ for states $s_i \in \mathbb{I}$, with $\mathbb{I}$ being the set of in-distribution data, i.e. the states encountered in training.
All possible states that were not encountered in training, i.e. $s_o \not\in \mathbb{I}$ define the set of out-of-distribution data $\mathbb{O}$.
If the policy entropy of all states in the in-distribution set is smaller than the entropy of all states in the out-of-distribution set:
\begin{equation}
    H(\pi(s_i)) < H(\pi(s_o)), \forall s_i \in \mathbb{I}, \forall s_o \in \mathbb{O}
\end{equation}
a decision boundary exists that allows for a perfect separation of in- and out-of-distribution states, making it possible to construct a perfect classifier with $tpr=1$, $fpr=0$ as described in section~\ref{eval_binary_clfs}.
In practice, the policy entropy distributions will most likely overlap, reducing the performance of a classifier constructed based on them.
In the following chapters, we present experiments conducted following the process described in section~\ref{subsec:training_testing_proc} to evaluate the performance, based on a reinforcement learning benchmark environment.

\section{A process for benchmarking OOD classification in reinforcement learning} \label{subsec:training_testing_proc}
\begin{figure}
    \includegraphics[width=\textwidth]{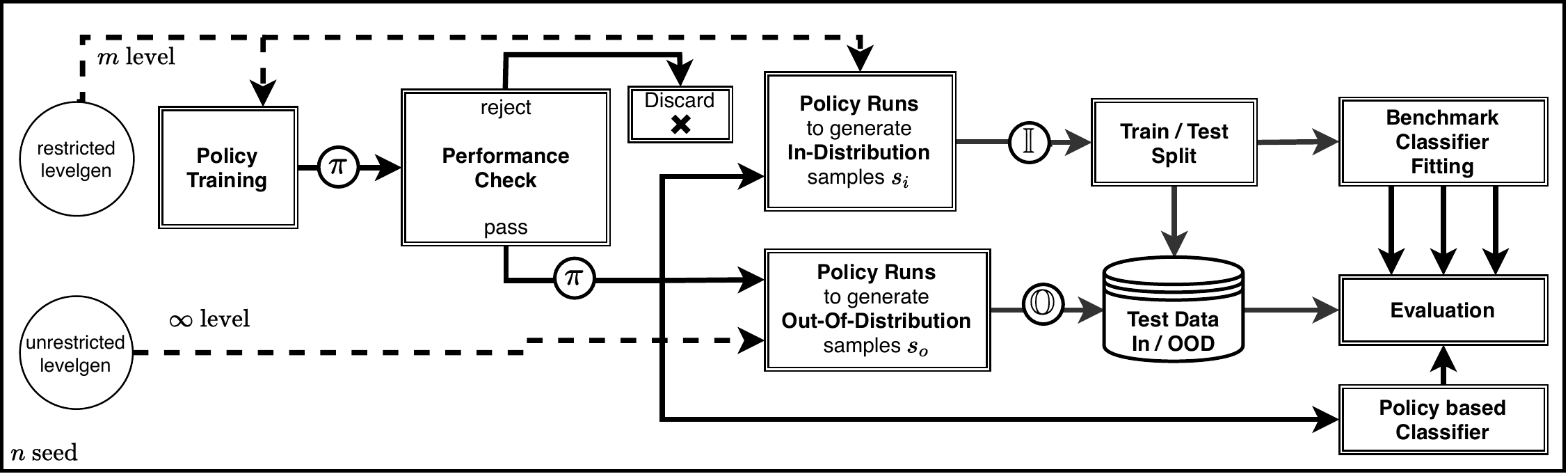}
    \caption{A process for benchmarking OOD classification in reinforcement learning. The complete process can be repeated $n$ times using different random seeds to average out variance in the classifier evaluation caused by random sampling.}
    \label{fig:process}
\end{figure}

In this section, we present a process for benchmarking out-of-distribution classification in reinforcement learning (Figure \ref{fig:process}).
This process encompasses a complete pipeline starting with the training of reinforcement learning policies, over in- and out-of-distribution state sample collection, (non-policy based) benchmark classifier fitting, leading up to the final classifier performance evaluations.\\
The complete process can be repeated ($n$ times) using different random seeds in order to average out variance in the classifier evaluation caused e.g. by random initialization of neural network weights or the level generators.
We call one such run a \textit{process-repeat}.
When running more than one process-repeat, it is possible to compute some central estimator of the performance (e.g. median and standard-deviation of the AUC) over the process-repeats, and visualize the performance results using e.g. box-plots.\\
For each process-repeat, \textit{policy training} is performed on a different set of $m$ levels for a fixed amount of timesteps.
As optimizing policy performance of the learning algorithm is not the focus, it is possible to perform a policy selection step after training, as would be done in a real-world usecase, where only the best policies get deployed in production.
The process continues only if the \textit{policy performance check} is passed, e.g. policy performance in training converged near the maximum return of the environment.
If not, the policy is \textit{discarded} and the next process-repeat continues from the beginning.\\
If the policy performance check was passed, multiple \textit{policy runs} are executed on the same set of $m$ levels as in training (called IND runs).
All states $s_i$ encountered during these IND runs are collected and together constitute the set of in-distribution states $\mathbb{I}$.
Separately, multiple \textit{policy runs} are executed using the unrestricted level generator, so that each policy run uses a new level from the generator (called OOD runs).
All states $s_o$ encountered during these runs are collected and together constitute the set of OOD states $\mathbb{O}$.
Figure~\ref{fig:coinrun_env} shows example states as generated during policy training, IND runs and OOD runs.\\
A \textit{train/test split} is performed on the set of collected in-distribution states $\mathbb{I}$.
Non-policy based \textit{benchmark classifiers} are fitted only on the train part of the in-distribution states, in order to prevent overfitting.
Policy based classifiers, like PEOC do not need a \textit{fitting} step on the in-distribution data, as they are based on the policy network learned during the policy training phase.
The test part of $\mathbb{I}$ is combined with the complete set $\mathbb{O}$ and constitutes the \textit{test data}, on which all classifiers are evaluated.
As for this evaluation, ground-truth labels (i.e. which set $\mathbb{I}$ or $\mathbb{O}$ a sample belongs to) are known, receiver operating characteristics of the classifiers can be calculated.\\
Note again, that this complete training \& classifier evaluation process, as described above, can be repeated $n$ times using different random seeds to compute some central estimators of the classifier performances.

\section{Experimental Setup}

\subsection{Environments}
Until recently, a lack of suitable benchmark environments made it difficult to evaluate out-of-distribution classification performance in deep reinforcement learning.
Standard evaluation environments for deep reinforcement learning like OpenAI Gym~\cite{gym} or the arcade learning environment~\cite{bellemare2013arcade} are not suitable, as it is necessary to create different in- and out-of-distribution state sets.
In the last two years, an increased research focus on generalization performance has lead to the development of new benchmark environments that allow for a separation of training- and test-environments~\cite{farebrother2018generalization,zhang2018study,cobbe2019procgen}.
Some of these environments are also suitable to evaluate out-of-distribution classification performance.

\begin{figure}
    \centering
    \begin{subfigure}{.32\textwidth}
        \includegraphics[width=.98\textwidth]{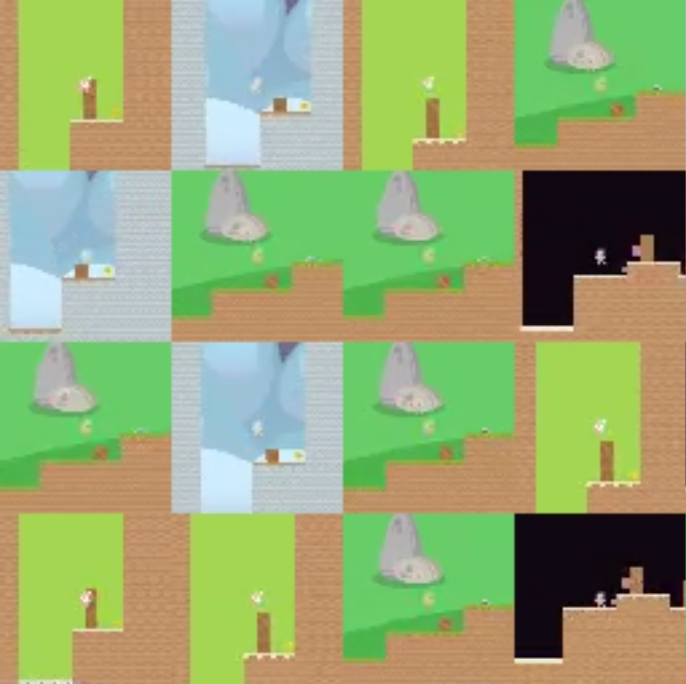}
        \caption{}
        \label{fig:coinrun_env_ind_l0_4x4}
    \end{subfigure}
    \begin{subfigure}{.32\textwidth}
        \includegraphics[width=.98\textwidth]{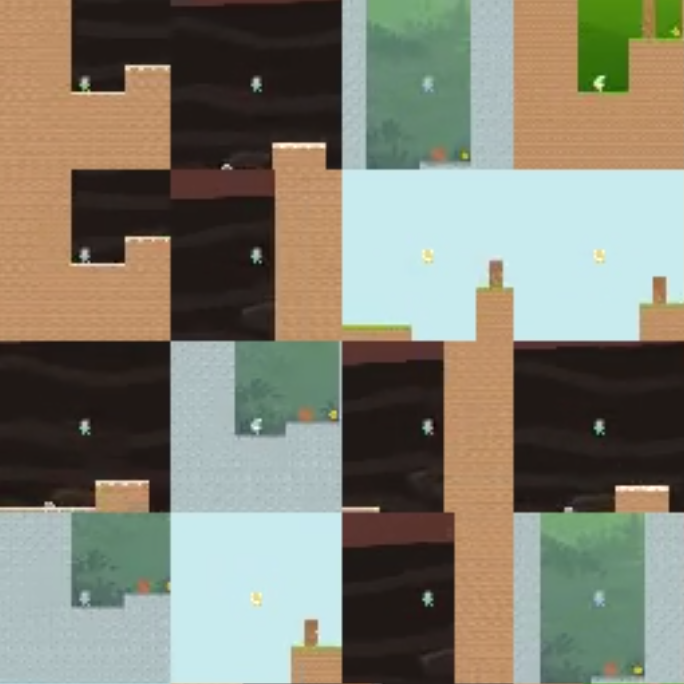}
        \caption{}
        \label{fig:coinrun_env_ind_l16_4x4}
    \end{subfigure}
    \begin{subfigure}{.32\textwidth}
        \includegraphics[width=.98\textwidth]{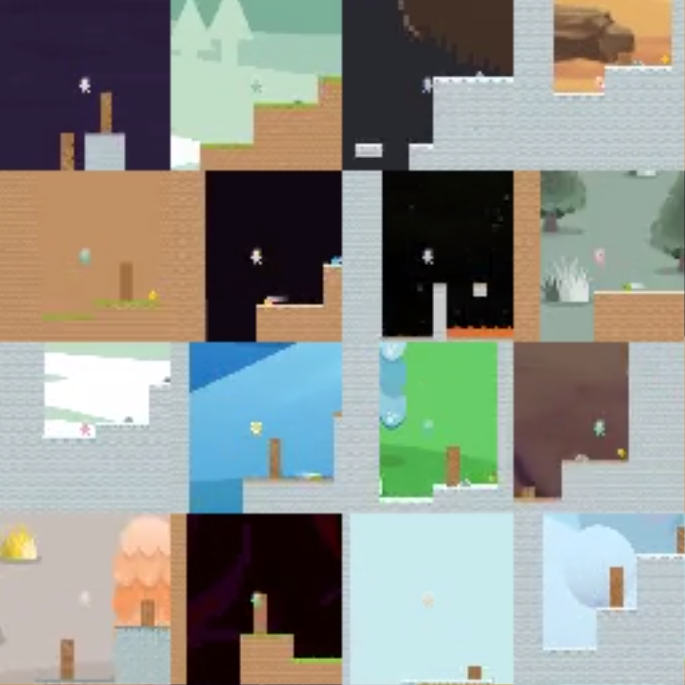}
        \caption{}
        \label{fig:coinrun_env_ood_4x4}
    \end{subfigure}
    \caption{Example states as generated during training/policy runs. Policy training and IND runs (\ref{fig:coinrun_env_ind_l0_4x4}, \ref{fig:coinrun_env_ind_l16_4x4}) are restricted to $4$ different levels, resulting in different, restricted sets of levels for each repeat. No seed restriction is applied to the level generator for OOD runs (\ref{fig:coinrun_env_ood_4x4}).}
    \label{fig:coinrun_env}
\end{figure}

For our experiments, we chose the CoinRun~\cite{cobbe2019procgen} environment as the level generator, as we deemed it's procedural level generation approach to be the most suitable to evaluate the proposed PEOC approach.
CoinRun is a simple platform jump\&run game, with discrete actions, where the agent has to collect the coin at the far right side of the level.
To do this, stationary obstacles, chasms and moving enemies have to be avoided, to avert death.
A state in the CoinRun environment is a color image encoded as a vector of size $64 \times 64 \times 3$.
Figure~\ref{fig:coinrun_env} shows example states of different levels as used for training and testing.

\subsection{Evaluation Algorithms and Hyperparameters}

We evaluate the performance of PEOC based on PPO2 as the RL algorithm.
For this, we make use of the PPO2 implementation from the OpenAI Baselines package~\cite{baselines}.
We combine PPO2 with an IMPALA Convolutional neural network architecture~\cite{impala}, as good results were achieved with this in related work.
The evaluated network consists of a convolutional sequence with depths $[16, 32, 32]$ followed by a max-pooling layer and 2 residual blocks.

We repeated the complete training \& classifier evaluation process as described in section~\ref{subsec:training_testing_proc} for $40$ times using different level seeds.
Each repeat, policy training was performed for \num{25e5} time steps.
We store a snapshot of the policy after the first policy update as well as after the last policy update and call the classifiers constructed based on them PEOC-1 and PEOC-150 respectively.
$8$ policies passed the performance check, i.e. the return converged at the achievable maximum of $10$.
Policy IND runs are executed for a total of {\raise.35ex\hbox{$\scriptstyle\sim$}}\num{30e3} steps, OOD runs for {\raise.35ex\hbox{$\scriptstyle\sim$}}\num{10e3} steps.
After performing a $2/1$ train/test split on the in-distribution set, this results in an evenly distributed test set containing {\raise.35ex\hbox{$\scriptstyle\sim$}}\num{20e3} in-and out-of-distribution samples on which the classifiers are evaluated.
Parameters are summarized in Table \ref{table:training_params}.

\begin{table}
    \caption{Training \& evaluation parameters}\label{tab1}
    \begin{center}
        \begin{tabular}{|c | c|}
            \hline
            \# training \& classifier evaluation process repeats & 40 \\
            \hline
            \# policies after perf. check & 8 \\
            \hline
            $m$ level per repeat & 4 \\
            \hline
            Policy training steps, per repeat & \num{25e5} \\
            \hline
            RL Policy selection & return converges at $10$ \\
            \hline
            IND run steps, per repeat & \num{30e3} \\
            \hline
            OOD run steps, per repeat & \num{10e3} \\
            \hline
            Classifier train/test split & $2/1$ \\
            \hline
        \end{tabular}
        \label{table:training_params}
    \end{center}
\end{table}

\noindent In order to benchmark the classification performance of PEOC, each repeat, $3$ non-policy based state-of-the-art classifiers are fit on the train split of the in-distribution data:
An autoencoder based approach, based on~\cite{aggarwal2015outlier} and the SO-GAAL and MO-GAAL approaches as presented in~\cite{liu2019generative}.
We use the implementation and default hyperparameters as provided by~\cite{zhao2019pyod} for all $3$ classifiers.

\section{Performance Results}

\begin{figure}
    \includegraphics[width=\textwidth]{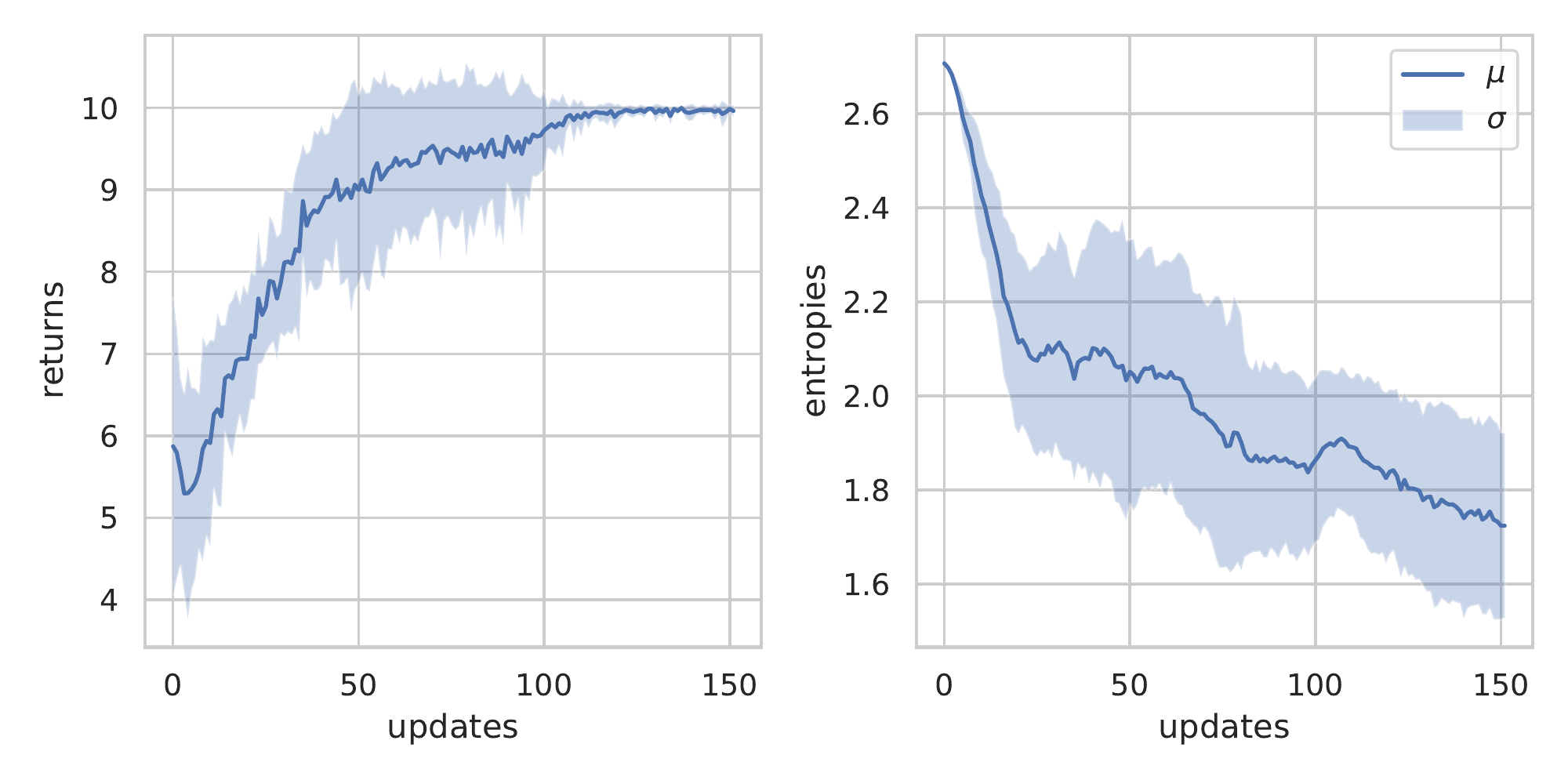}
    \caption{Reward and policy entropy over training updates of the $8$ policies that passed the performance check.
    The solid line and shaded regions represent the mean $\mu$ and standard deviation $\sigma$ of the successful training runs, respectively.}
    \label{fig:train_avg}
\end{figure}

Of the $40$ process-repeats executed, $8$ policies passed the policy performance check after \num{25e5} training steps,
i.e. performance of the respective policy converged at $10$, the maximum achievable return of the CoinRun environment.
Figure~\ref{fig:train_avg} left shows mean and standard-deviation of the achieved return of these policies against the number of training updates.
As expected, return performance increases, reflecting the discovery of increasingly successful policies.
Policy entropy (Figure~\ref{fig:train_avg} right) decreases over training progress, confirming the hypothesis that the entropy of the action distribution has to decrease for states encountered during training in order to act optimally.

\begin{figure}
    \centering
    \begin{subfigure}{.49\textwidth}
        \includegraphics[width=.98\textwidth]{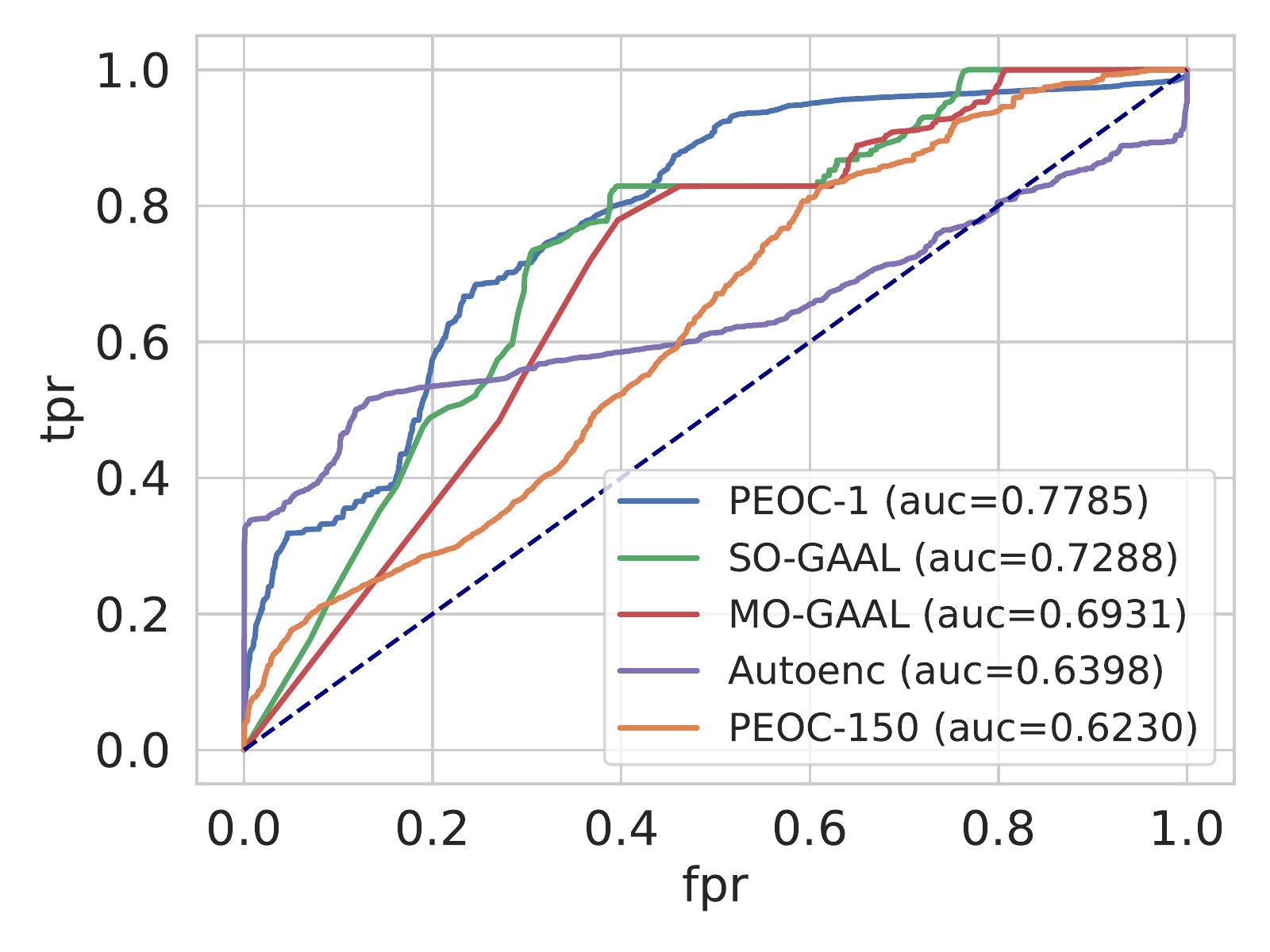}
        \caption{}
    \end{subfigure}
    \begin{subfigure}{.49\textwidth}
        \includegraphics[width=.98\textwidth]{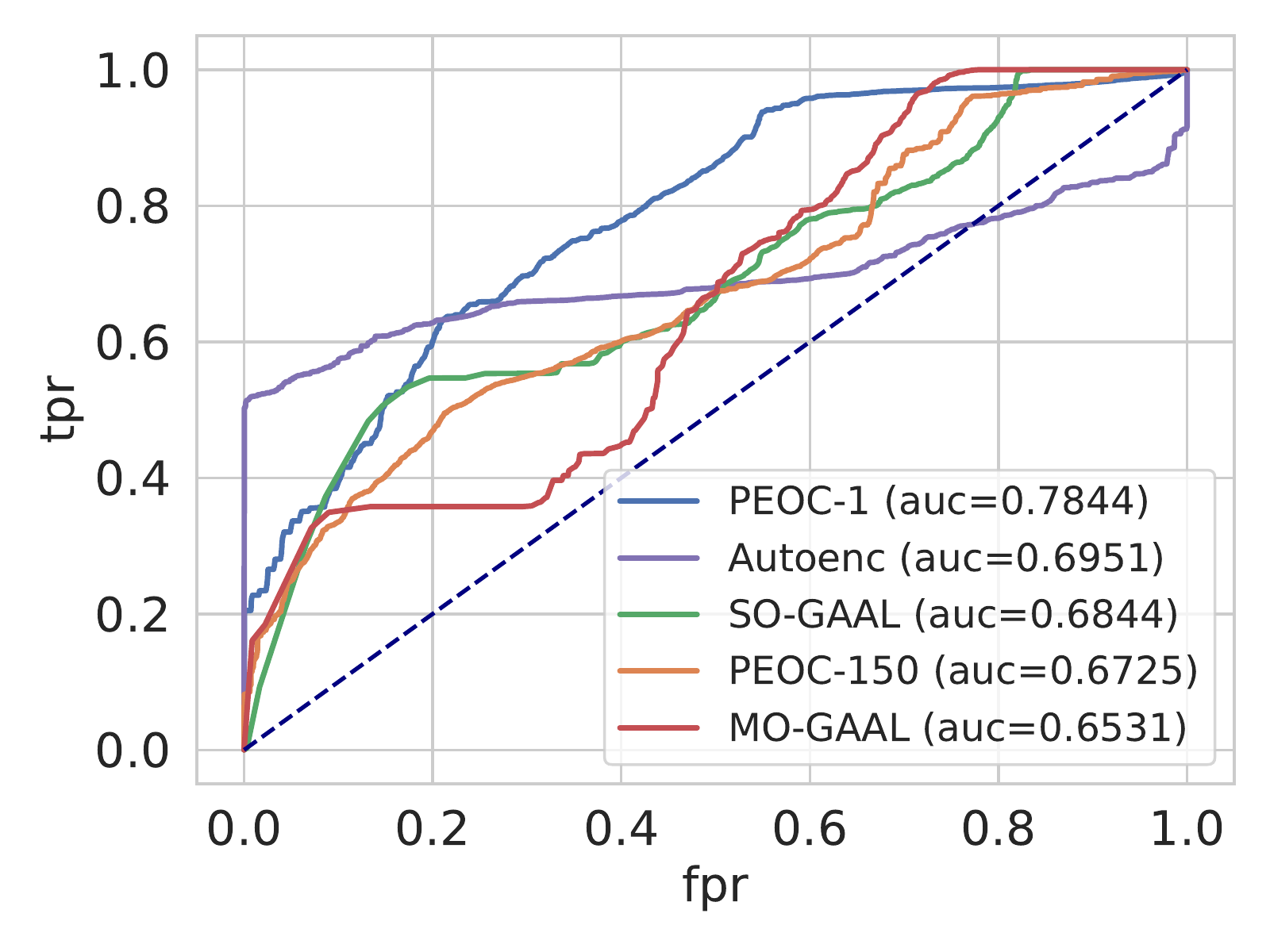}
        \caption{}
        \label{fig:roc_b}
    \end{subfigure}
    \begin{subfigure}{.49\textwidth}
        \includegraphics[width=.98\textwidth]{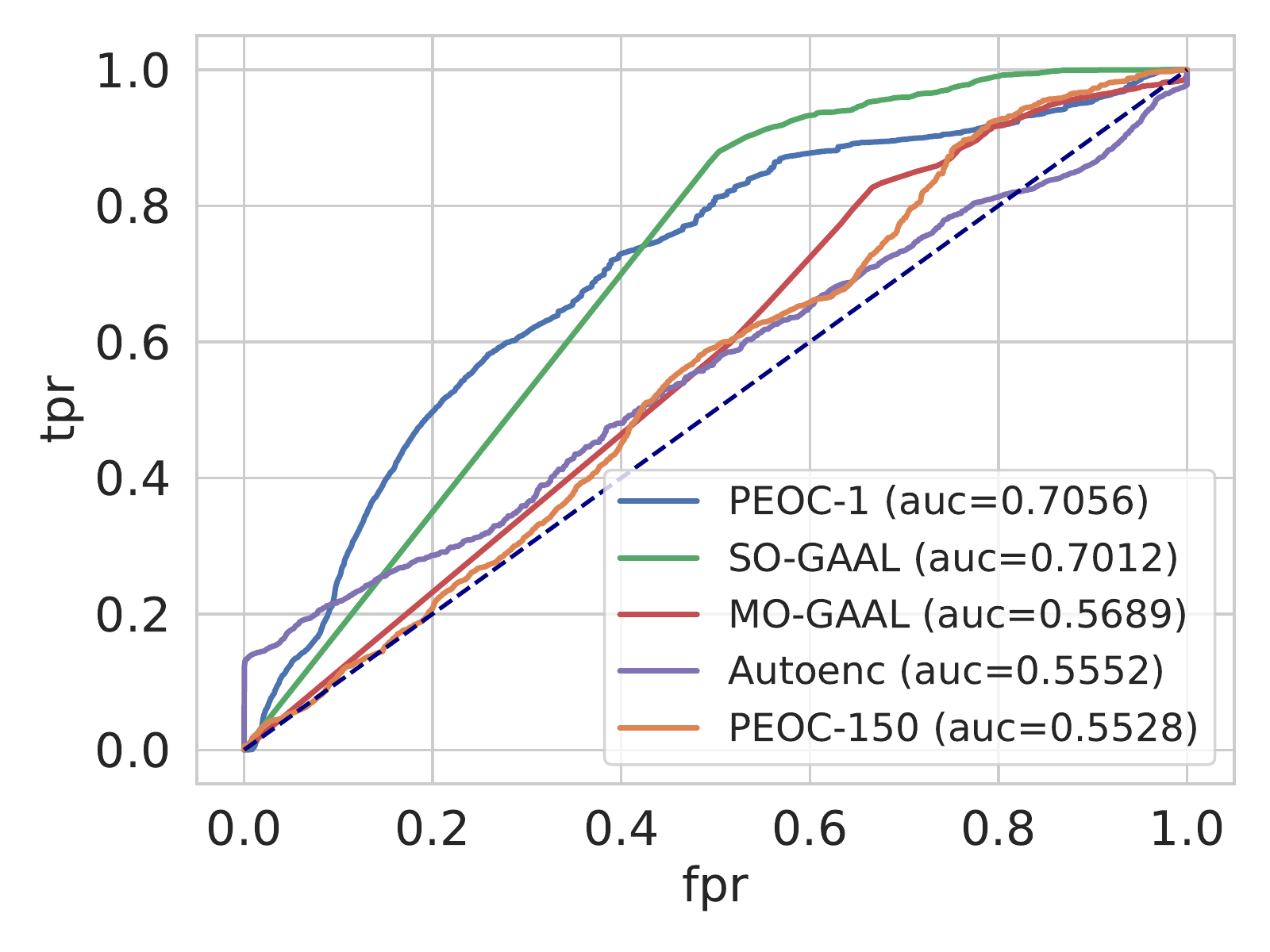}
        \caption{}
    \end{subfigure}
    \begin{subfigure}{.49\textwidth}
        \includegraphics[width=.98\textwidth]{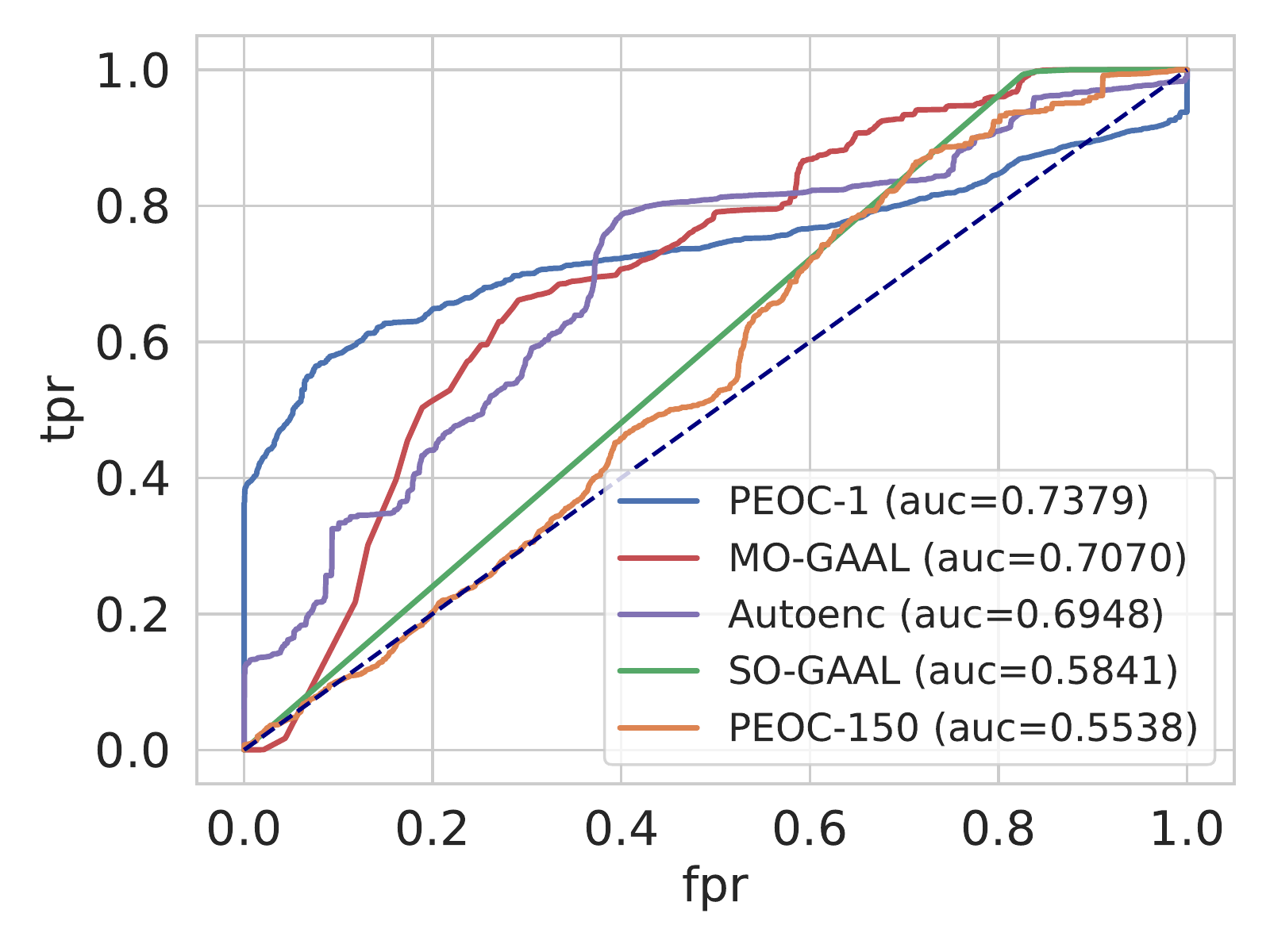}
        \caption{}
    \end{subfigure}
    \begin{subfigure}{.49\textwidth}
        \includegraphics[width=.98\textwidth]{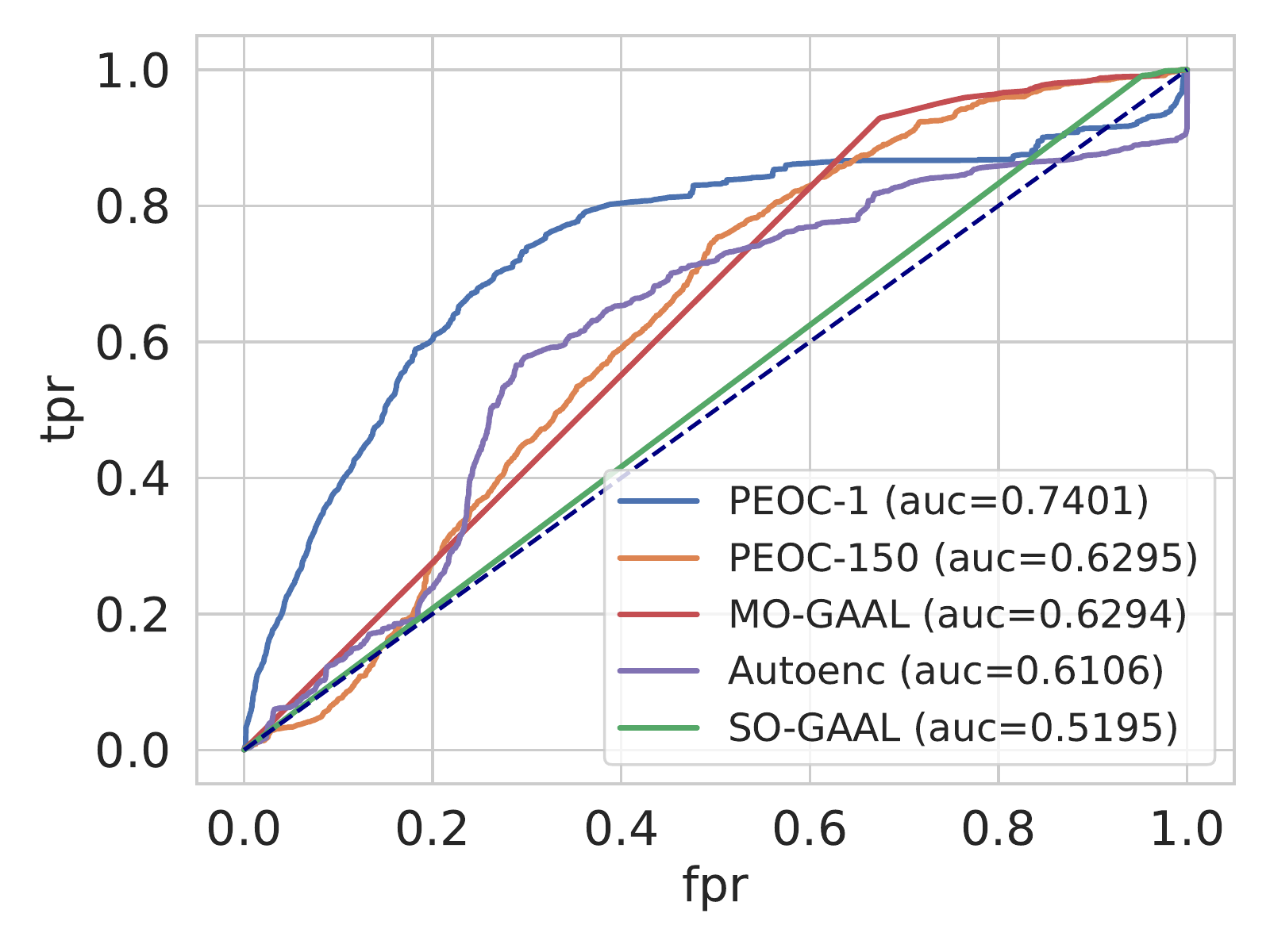}
        \caption{}
    \end{subfigure}
    \begin{subfigure}{.49\textwidth}
        \includegraphics[width=.98\textwidth]{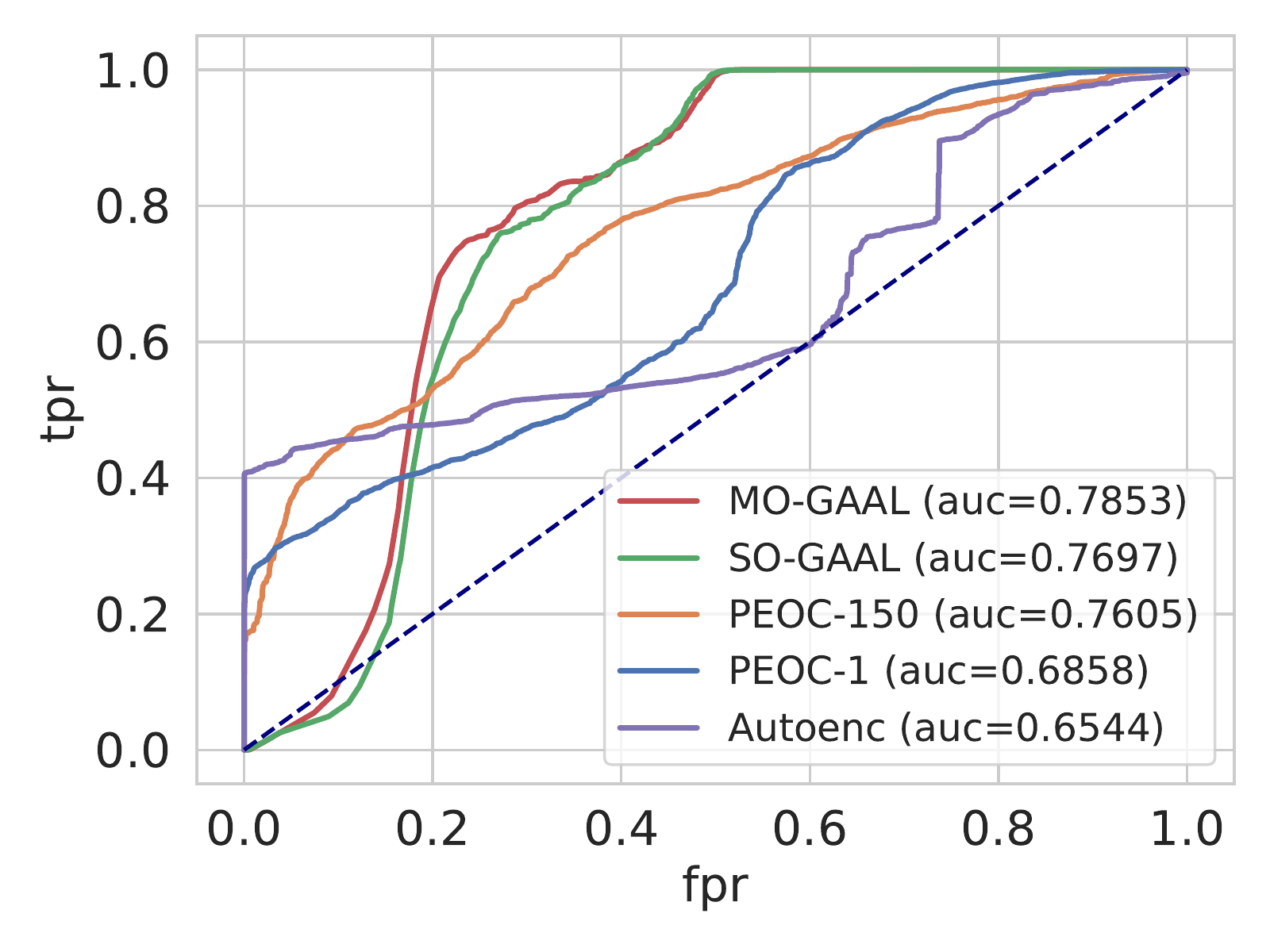}
        \caption{}
        \label{fig:roc_f}
    \end{subfigure}
    \begin{subfigure}{.49\textwidth}
        \includegraphics[width=.98\textwidth]{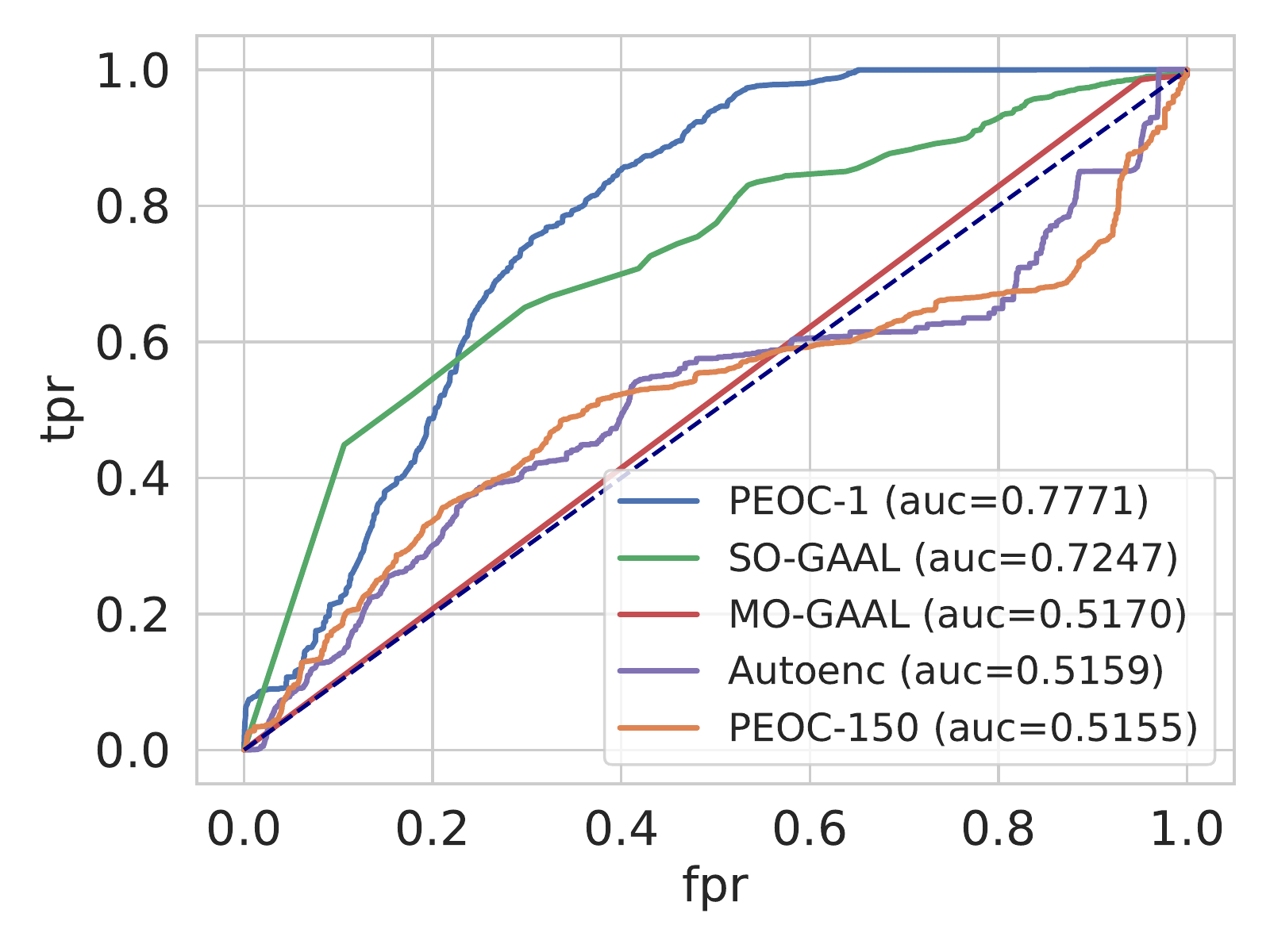}
        \caption{}
    \end{subfigure}
    \begin{subfigure}{.49\textwidth}
        \includegraphics[width=.98\textwidth]{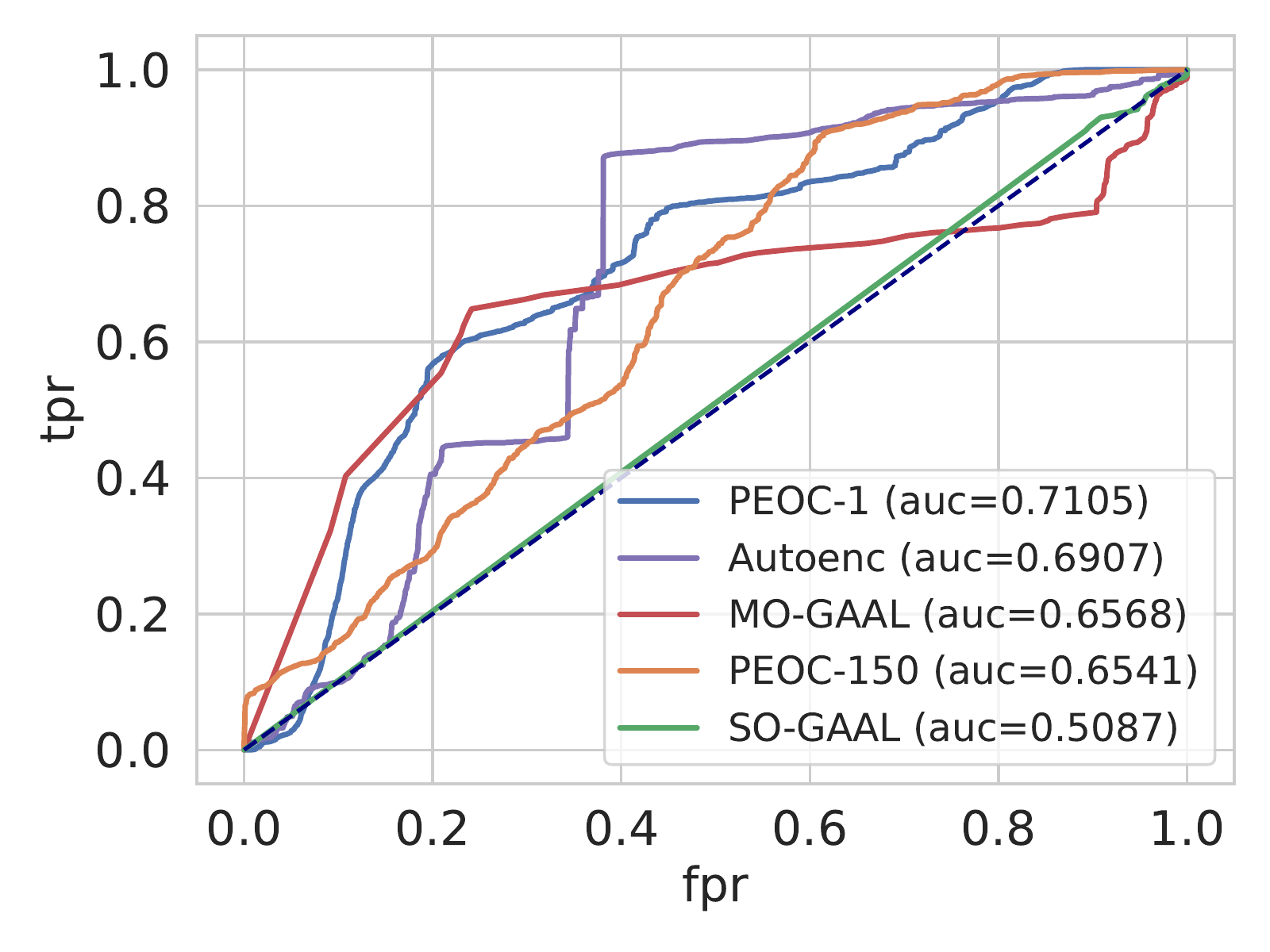}
        \caption{}
    \end{subfigure}
    \caption{ROC plots of the classifier evaluations of the $8$ process-repeats. Each plot a)-h) shows the true-positive rate ($tpr$) against the false-positive rate ($fpr$) achieved by the classifier on the respective test data set.}
    \label{fig:roc}
\end{figure}

Each of these policies was evaluated for \num{30e3} steps on the respective environments they were trained on, collecting $8$ separate sets of in-distribution samples.
For each of these, separate instances of the $3$ benchmark classifiers were then fit on the train split of the respective in-distribution set, while testing of all classifiers was performed on the respective OOD set.
The classification scores were then visualized in $8$ separate ROC plots to compare the performance of the respective benchmark classifiers.
In addition, the area under the curve (ROC AUC) was computed.
Figure~\ref{fig:roc} shows that none of the classifiers reaches perfect classification results in any of the process-repeats.
Still, some classifiers perform better than others. PEOC-1, i.e. the policy entropy based out-of-distribution classifier using the policy snapshot after the first update performs best across all process-repeats, when considering the area under the curve (AUC) with values ranging from $0.7056$ to $0.7844$. Even so, there are exceptions as can be seen in Figure~\ref{fig:roc_f}, where the MO-GAAL classifier achieved the highest AUC of $0.7853$. Apart from the raw AUC values, it becomes apparent from the ROC curves, that for some evaluations (e.g. AE in Figure~\ref{fig:roc_b}), perfect $fpr=0$ can be achieved while still classifying more than $40\%$ of the OOD samples correctly.
PEOC-150 mostly shows rather low performance, underperforming the other classifiers for most process-repeats.
With an AUC varying between $0.5155$ and $0.7605$ the performance is not reliable in summary.

\begin{figure}
    \centering
    \includegraphics[width=0.5\textwidth]{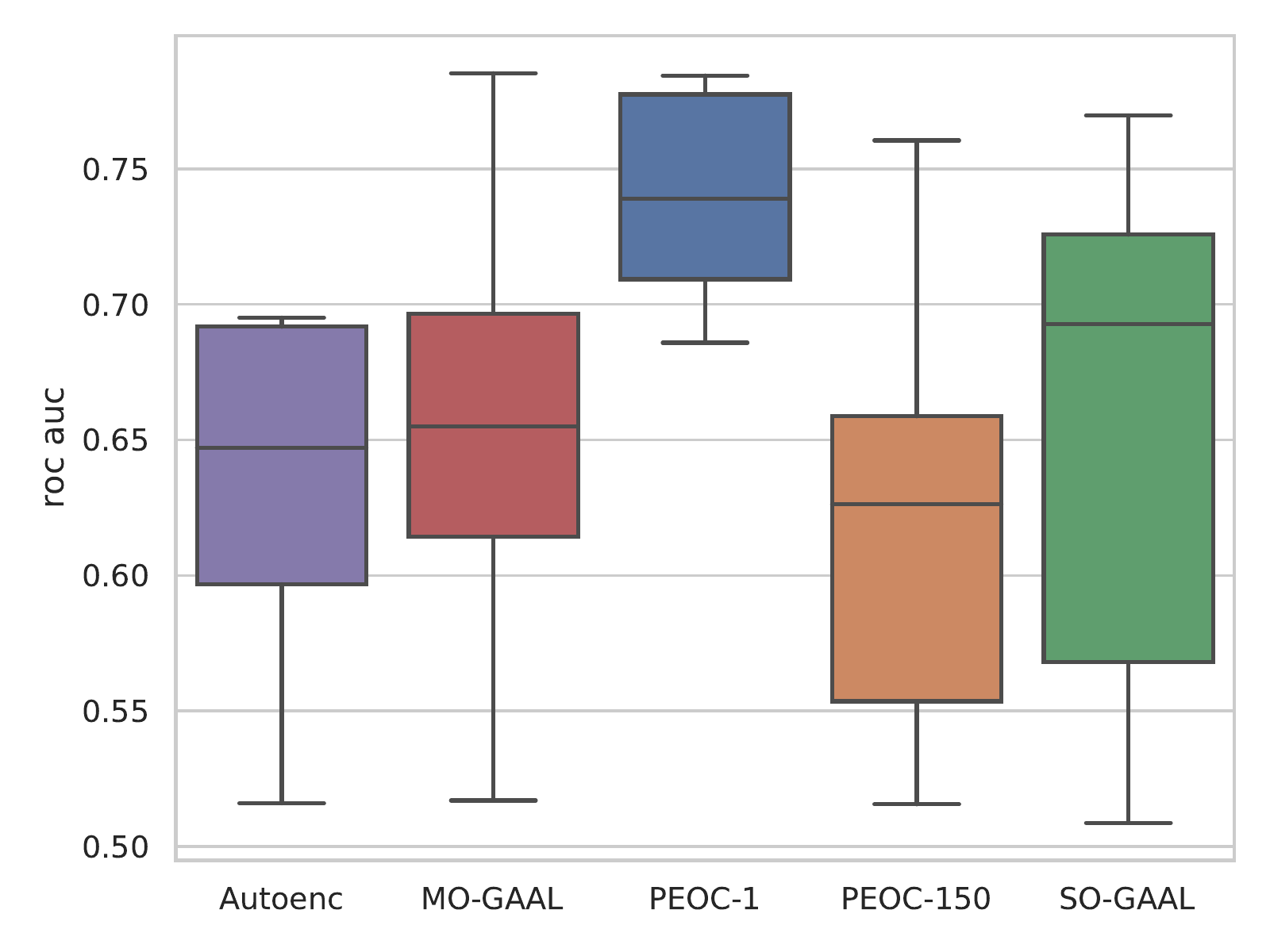}
    \caption{Comparison of classifier performance based on ROC AUC values of $8$ process-repeats.}
    \label{fig:auc}
\end{figure}

Using the ROC AUC values of all process-repeats, we compare the overall performance of the different classifiers, i.e. we summarize the $8$ receiver operating characteristics for each classifier.
Figure~\ref{fig:auc} shows this overall classifier performance in the form of a box-plot.
Using the median ROC AUC as a central measures of classifier performance, PEOC-1 surpases the other classifiers with a value of $0.74$. PEOC-150 shows a far lower median value of $0.63$.
The non-policy based benchmark classifiers' median ROC AUC are $0.65$ for the autoencoder, $0.69$ for SO-GAAL and $0.65$ for MO-GAAL.
\vspace{-0.8cm}
\section{Discussion and Future Work}
In this work, we proposed PEOC, a policy entropy based out-of-distribution classifier as well as a structured process for benchmarking OOD classification in RL that can be reused for comparing different OOD classification approaches in the future.
Performance evaluation results on the procedural CoinRun~\cite{cobbe2019procgen} environment show that PEOC-1 (using the policy after the first update) based on PPO2 as the RL algorithm, is highly competitive against state-of-the-art one-class classification algorithms,
i.e. it reliably classifies out-of-distribution states.

Interestingly, using the final policy as the basis of the classifier did show inferior performance when compared to the policy after the first update.
As to why this is the case, one hypothesis is that the convolutional layers of the policy network at first learn general features representing the states of the environment which seem to work well to differentiate in- from out-of-distribution states.
With further progressing training, the network then concentrates on features relevant to policy performance optimization which might be less relevant for OOD detection.
We aim to analyse this further using visualization approaches from the field of interpretable machine learning.

As our current evaluation was limited to only one policy-gradient based approach, i.e. PPO2, it will be interesting to see if the choice of RL algorithm has an influence on classifier performance.
Another interesting question is the behaviour of PEOC when using a policy that successfully generalizes to unencountered states.
Being able to differentiate between states seen in training, states the policy generalizes to and completely out-of-distribution states, where no generalization is possible, would be extremely valuable for deploying RL agents in the real world.

%
%
%
\bibliographystyle{splncs04}
%
\newpage
\bibliography{main}

\begin{thebibliography}{10}
\providecommand{\url}[1]{\texttt{#1}}
\providecommand{\urlprefix}{URL }
\providecommand{\doi}[1]{https://doi.org/#1}

\bibitem{aggarwal2015outlier}
Aggarwal, C.C.: Outlier analysis. In: Data mining. pp. 237--263. Springer
  (2015)

\bibitem{andrychowicz2020learning}
Andrychowicz, O.M., Baker, et~al.: Learning dexterous in-hand manipulation. The
  International Journal of Robotics Research  \textbf{39}(1),  3--20 (2020)

\bibitem{bellemare2013arcade}
Bellemare, M.G., Naddaf, Y., Veness, J., Bowling, M.: The arcade learning
  environment: An evaluation platform for general agents. Journal of Artificial
  Intelligence Research  \textbf{47},  253--279 (2013)

\bibitem{berner2019dota}
Berner, Christopher, B., et~al.: Dota 2 with large scale deep reinforcement
  learning. arXiv preprint arXiv:1912.06680  (2019)

\bibitem{gym}
Brockman, G., Cheung, V., Pettersson, L., Schneider, J., Schulman, J., Tang,
  J., Zaremba, W.: Openai gym (2016)

\bibitem{cobbe2019procgen}
Cobbe, K., Hesse, C., Hilton, J., Schulman, J.: Leveraging procedural
  generation to benchmark reinforcement learning. arXiv preprint
  arXiv:1912.01588  (2019)

\bibitem{baselines}
Dhariwal, P., Hesse, C., Klimov, O., Nichol, A., Plappert, M., Radford, A.,
  Schulman, J., Sidor, S., Wu, Y., Zhokhov, P.: Openai baselines (2017)

\bibitem{impala}
Espeholt, L., Soyer, H., Munos, R., et~al.: {IMPALA:} scalable distributed
  deep-rl with importance weighted actor-learner architectures. CoRR  (2018)

\bibitem{farebrother2018generalization}
Farebrother, J., Machado, M.C., Bowling, M.: Generalization and regularization
  in dqn (2018)

\bibitem{haarnoja2017reinforcement}
Haarnoja, T., Tang, H., Abbeel, P., Levine, S.: Reinforcement learning with
  deep energy-based policies. In: Proceedings of the 34th International
  Conference on Machine Learning-Volume 70. pp. 1352--1361. JMLR. org (2017)

\bibitem{hendrycks16}
{Hendrycks}, D., {Gimpel}, K.: {A Baseline for Detecting Misclassified and
  Out-of-Distribution Examples in Neural Networks}. ArXiv e-prints  (Oct 2016)

\bibitem{liang17}
{Liang}, S., {Li}, Y., {Srikant}, R.: {Enhancing The Reliability of
  Out-of-distribution Image Detection in Neural Networks}. ArXiv e-prints  (Jun
  2017)

\bibitem{liu2019generative}
Liu, Y., Li, Z., Zhou, C., Jiang, Y., Sun, J., Wang, M., He, X.: Generative
  adversarial active learning for unsupervised outlier detection. IEEE
  Transactions on Knowledge and Data Engineering  (2019)

\bibitem{mnih2016asynchronous}
Mnih, V., Badia, A.P., Mirza, M., Graves, A., Lillicrap, T.P., Harley, T.,
  Silver, D., Kavukcuoglu, K.: Asynchronous methods for deep reinforcement
  learning. CoRR  (2016)

\bibitem{pimentel14}
Pimentel, M.A., Clifton, D.A., Clifton, L., Tarassenko, L.: A review of novelty
  detection. Signal Processing  \textbf{99},  215 -- 249 (2014)

\bibitem{puterman2014markov}
Puterman, M.L.: Markov decision processes: discrete stochastic dynamic
  programming. John Wiley \& Sons (2014)

\bibitem{schulman2017proximal}
Schulman, J., Wolski, F., et~al.: Proximal policy optimization algorithms. CoRR
   (2017)

\bibitem{ubood}
Sedlmeier., A., Gabor., T., Phan., T., Belzner., L., Linnhoff{-}Popien., C.:
  Uncertainty-based out-of-distribution classification in deep reinforcement
  learning pp. 522--529 (2020)

\bibitem{sutton1998introduction}
Sutton, R.S., Barto, A.G.: Introduction to reinforcement learning, vol.~135.
  MIT press Cambridge (1998)

\bibitem{williams1991function}
Williams, R.J., Peng, J.: Function optimization using connectionist
  reinforcement learning algorithms. Connection Science  \textbf{3}(3),
  241--268 (1991)

\bibitem{zhang2018study}
Zhang, C., Vinyals, O., Munos, R., Bengio, S.: A study on overfitting in deep
  reinforcement learning (2018)

\bibitem{zhao2019pyod}
Zhao, Y., Nasrullah, Z., Li, Z.: Pyod: A python toolbox for scalable outlier
  detection. Journal of Machine Learning Research  \textbf{20}(96), ~1--7
  (2019)

\end{thebibliography}
\end{document}